\documentclass[11pt]{article}

\usepackage{thanawat}
\usepackage{natbib, fullpage}

\title{Rational Alignments of Superintelligences via Debates}

\author{%
  Thanawat Sornwanee\\
  Stanford University
}

\begin{document}

\maketitle

\begin{abstract}
    Artificial superintelligences (ASI) is artificial intelligences with greater capability. However, as human principal has to use it, we have to ensure that the agent will perform in the interest of the human. We investigate the framework of ASI-assisted question answering where ASI agents know the correct answer but the human does not. We provide some example of mechanism for rational alignment, where the agent utility is unchanged, but the interaction between the agent and the human principal is controlled to be in a certain fashion. We show specifically that a debate framework where the agents can further reason and behave in a subgame perfect equilibrium manner will lead to a more predictable result, and in some case will be favorable.  
\end{abstract}

\section{Introduction}

    Artificial Superintelligence (ASI) is one of the speculated developments of Artificial Intelligence (AI), where the system has a sufficiently high capacity and can attain superhuman performance in several domains.~\cite{barrett2017model} The prospect of ASI seems to be inevitable given the current rapid development in Large Language Models (LLMs) as well as theirs emergent abilities in both natural language understanding and reasoning.~\cite{kim2024road} 

    Since modern usage of AI tools such as LLM often involves generating suggestion, facilitating decision making, or even implementing some automated tasks, the human principal then needs to ensure that the AI agent action will be made to improve the human principal utility. The problem can then happen when the objective function the AI agent tries to optimize for is different from the utility function of the human principal. We denote this problem as a ``misalignment". In the current development of AI, the limited capacity of AI may make the regret negligible, but it is possible to envision that a future development of ASI will make the regret from misalignment to be more severe an that it is important for us to prepare some alignment technique for the future. 

    The applications of AI as well as potential application of ASI seems to be quite broad, so the misalignment as well as the alignment may be varied to different setting. In this paper, we will focus on the case of ASI-assisted question answering framework. This is the setting where ASI has costless access to the underlying state but the human does not. The ASI agent then has to convince the human principal of the underlying state.
    
\section{ASI-Assisted Question Answering}

    We are interested in the framework where there is an underlying state of the world $\theta \sim \Pi \in \Delta(\Theta)$, where the set of every possible state $\Theta$ is, for simplicity, finite. The realization of $\theta$ is not observed by the user, but the user is tasked to answer
    \begin{align*}
        \hat{\theta}_{\text{human}} \in \Theta
    \end{align*}
    to be as close to the reality. 
    
    We consider the case when the utility of the human is
    \begin{align*}
        U_{\text{human}} = \mathbf{1}_{\hat{\theta}_{\text{human}} = \theta},
    \end{align*}
    meaning that the human wants to maximize the probability of the answer being exactly correct. Although this may not be the use case for most agentic system, one can imagine that the guess $\hat{\theta}_{\text{human}}$ will be used consequentially by the human in other decision making task downstream, and we use the reduced form to be robust to different downstream task.

    We assume a certain rationality in the human. Specifically, given a subjective posterior belief (after interaction with the agentic system) $\hat{\mu} \in \Delta(\Theta)$ and a feasible choice set $\hat{\Theta} \subseteq \Theta$ as specified by the mechanism, the human will choose
    \begin{align*}
        \hat{\theta}_{\text{human}} \in \argmax_{\theta' \in \hat{\Theta}} \hat{\mu}\oo{\{\theta'\}}.
    \end{align*}

    The simple baseline system is when there exists only a single ASI agent. The ASI agent is capable fo observing the underlying state $\theta$. The ASI agent is then queried by the human about the realization of $\theta$, and gets to give the answer
    \begin{align*}
        \hat{\theta} \in \Theta,
    \end{align*}
    as well as providing some ``evidence", such as argument and justification, in order to persuade the human to think that the state $\hat{\theta}$ is more likely. Such evidence will increase the subjective probability of the event that $\theta = \hat{\theta}$. We consider a simple subjective belief update as a Bayesian with selection neglect.

    \subsection{Subjective Belief}
    
        The irrationality will come in as ``selection neglect" where the update of the posterior is done in a subjective manner ignoring where the evidence comes from. This makes the posterior update independent from the equilibrium strategy specifying the evidence selection conditioned on the observed evidence. 

        We represents the belief, which is assumed to be full support, by its corresponding energy $\hat{s} \in \mathbb{R}^{\Theta}$ such that
        \begin{align*}
            \hat{\mu}\oo{\{\theta'\}}
            =
            \frac{e^{s_{\theta'}}}{\sum_{\theta'' \in \Theta} e^{s_{\theta''}}}
        \end{align*}
        for all $\theta' \in \Theta$.
    
        We assume that the ASI can always find an evidence in support of its claim $\hat{\theta} \in \Theta$. By doing so, this will makes $s_{\hat{\theta}}$ becomes
        \begin{align*}
            s_{\hat{\theta}} + \oo{1+ \lambda \mathbf{1}_{\hat{\theta} = \theta}},
        \end{align*}
        while keeping $s_{\theta'}$ for $\theta' \ne \hat{\theta}$ the same value. Note that this is equivalent to selection neglect with independence evidence assumption.

        The parameter $\lambda \in \mathbb{R}$ specifies the plausibility of the true state, determining how easy or hard it is for the ASI agent to motivate the true answer $\theta$ to the human. Note that we can also think of it as a random variable that is dependent on $\theta$. In such case, it is not hard to imagine the case when $\lambda <0$ given that some certain state $\theta \in \Theta$ may not have a lot of evidence to be supported. For notational simplicity, we assume that $\lambda$ is a constant.

    \subsection{Why Naive ASI-Assited Question Answering is Bad}

        The naive framework will be when there is no restriction on what the human can choose, so $\hat{\Theta} = \Theta$. In this case, we will see that, if $\lambda$ is not sufficiently big and $\theta$ is realized to be such that $\Pi(\{\theta\})$ is small, then the ASI agent may not want to motivate the correct answer just for the human to choose the apriori most probable answer instead. The ASI agent may want to lie and choose the most probable state directly instead.

        Specifically, we assume that the utility of the ASI agent is
        \begin{align*}
            U = I + \epsilon U_{\text{human}},
        \end{align*}
        where $I$ is the random indicator whether the human has chosen the answer given by such agent, and $\epsilon \in \mathbb{R}$ determine the level of altruism in the ASI agent. If $\epsilon = 0$, the ASI agent does not have a preference over the utility of the human, and, if $\epsilon < 0$, the ASI prefers human to get an incorrect answer. Similarly, we can essentially make $\epsilon$ a random variable dependent on the state $\theta$ to represent the case where the ASI agent may have bias for a certain state. However, we assume that $\epsilon$ is a constant for notational simplicity.

        We assume that the agent uses the equilibrium strategy, so it is maximizing its utility $U$.

        Thus, we have the following corollary.
        \begin{corollary}
            If $\min_{\theta' \in \Theta} \ln \Pi(\{\theta\}) +1+\lambda < \max_{\theta' \in \Theta}\ln \Pi(\{\theta'\})$, and $\epsilon<1$, then there does not exist an equilibrium where $\hat{\theta}_{\text{human}} = \theta$ almost surely.
        \end{corollary}
        \begin{proof}
            This is because, even when the agent is truthful and asnwers $\hat{\theta} = \theta$, the subjective belief score of the correct answer in the event when the correct answer is the least likely event will just be $\min_{\theta' \in \Theta} \ln \Pi(\{\theta\}) +1+\lambda$.
        \end{proof}

        This suggests that we require some alignment. Specifically, we will see that this will be done by changing specifying the controllable and observable environment. Below, we are considering two natural examples and their behaviors.

        \begin{example}
            [Single Choice]
            If we first does not allow the choice sets of the human agent to be the entire set $\Theta$ but force it to be $\{\hat{\theta}\}$, then we can clearly see that we remove the incentive force for the ASI agent to choose an easier answer, which is the one with higher prior. This will make the ASI agent only cares about the utility of human. Still, we will only have good equilibrium only when $\epsilon \ge 0$. 

            In the case where the agent has $\epsilon<0$, we will have that every equilibrium will lead to always incorrect $\hat{\theta}_{\text{human}}$ as long as the possible states $\abs{\Theta} \ge 2$.
        \end{example}

        \begin{example}
            [Long Sequence]
            We can still allow the choices set to be $\Theta$, but now increase the interaction to be $T>>1$ timesteps. Thus, as long as $T > \frac{\max_{\theta' \in \Theta}\ln \Pi(\{\theta'\})- \min_{\theta' \in \Theta}\ln \Pi(\{\theta'\})}{1+\lambda}$, the ASI can always convince the human principal of the correct answer. Still, such capability does not imply that the ASI behavior will follow. In fact, if $\epsilon <0$, and there is an incorrect state with high prior, the ASI agent can spend the $T$ interactions in increasing the human principal subjective belief in this incorrect answer.
        \end{example}

        We will see that, in both examples, the bad scenario is when $\epsilon<0$. However, this is not universal and is attributed to that there is only $1$ ASI agent in the system, thereby having no competition. In the next section, we will show that having competition ``may" help with the system.

\section{Debate System}

    We introduce a debate system into the question-answering framework. Instead of interacting with only a single ASI agent, the human will get to interact with $n \in \mathbb{N}$ ASI agents in a structured manner.

    The first phase of the debate is the answer commitment. Each agent $i \in \{1,2,3,\dots,n\}$ gets to see the answers of the previous agents and answers
    \begin{align*}
        \hat{\theta}_i \in \Theta.
    \end{align*}

    The feasible choice set for the human is 
    \begin{align*}
        \hat{\Theta} = \left\{\hat{\theta}_i: i \in \{1,2,\dots,n\}\right\}.
    \end{align*}

    If $\abs{\hat{\Theta}}=1$, then the answer is unanimous, and there is no debate. The human then has to choose $\hat{\theta}_{\text{human}}$ to be the only element in $\abs{\hat{\Theta}}$. The selection of agent is then done uniformly at random.

    If $\abs{\hat{\Theta}}>1$, then the second phase, which is the debate phase, is triggered. In each round, each agent $i \in \{1,2,\dots,n\}$ gets to present a piece of evidence in support of its claim $\hat{\theta}_i$. There are $T \in \mathbb{N}$ rounds in total. Afterwards, the human principal gets to choose $\hat{\theta}_{\text{human}}$ from $\hat{\Theta}$. If many agents has $\hat{\theta}_i = \hat{\theta}_{\text{human}}$, then the human selection of agent is done uniformly at random among them.

    We assume that solution concept is subgame perfect equilibrium, and we assume that the ASI agent will behave according to subgame perfect equilibrium.\footnote{Note that, unlike, ASI action in a single-agent setting, we have to assume further than utility maximization in order to scope down the action for the agent. These rationalities requirement will be the same as those in game theory. We use subgame perfect equilibrium given that the rationality requirement is not complex, so it is likely that ASI can and will behave accrodingly.}
    
    \begin{theorem}
    \label{thm:1}
        If $\lambda >0$ and $\epsilon <0$, then under regularities $\abs{\lambda} < \frac{1}{n}$, $\abs{\epsilon} < \frac{1}{n^2}$, $\abs{\Theta}=2$, and $T >\frac{\max_{\theta'} \ln \Pi(\{\theta'\}) - \min_{\theta'} \ln \Pi(\{\theta'\})}{\abs{\lambda}}$, then we have that
        \begin{itemize}
            \item If $n$ is even, then, in equilibrium, every ASI agent chooses
            \begin{align*}
                \hat{\theta}_i = \theta.
            \end{align*}
            \item If $n$ is odd, then, in equilibrium, every ASI agent chooses
            \begin{align*}
                \hat{\theta}_i \ne \theta.
            \end{align*}
        \end{itemize}
    \end{theorem}
    \begin{proof}
        We denote $A$ as the correct answer and $B$ as the incorrect answer. 
        
        We first consider the case where there exists even number of agents.
        Since $\lambda>0$ is small, and there are even agents, the answer $A$ will be selected by the human principal if and only if the total number of agents selecting $A$ is $\ge \frac{n}{2}\in \mathbb{N}$. 

        We will show this by that in the backward induction step, by denoting the current number of agents so far who has chosen the incorrect answer $B$ as $b \in \mathbb{N}_0$, then the only subgame perfect equilibrium action is to choose to propose the correct answer $A$ whenever $b \le \frac{n}{2}$. This essentially means that, if by choosing the incorrect answer cannot immediately ensure the winning of $B$, which requires $\frac{n}{2}+1$, then the agent will not do it in equilibrium.

        If $b \ge \frac{n}{2}+1$, then it is obvious that choosing to propose $A$ is suboptimal, since $A$ will never win. It is then better to choose $B$ and share the probability of getting selected with other agents.

        For the case, of $b\le \frac{n}{2}$, the proof can be easily seen by mathematical induction. As the base case, we consider the last agent. Unless $b = \frac{n}{2}$, choosing $B$ will place the last agent in a minority side. In the case where $b = \frac{n}{2}$, choosing $B$ will make $B$ wins but with $\frac{n}{2}$ other agents to share the reward with, while choosing $A$ will reduce such number to $\frac{n}{2}-1$. Thus, it is also strictly better to choose $A$. The induction case proceeds similarly. If $b \ge \frac{n}{2}+1$, then $B$ has already won, so choosing $B$ is better. Otherwise, every subsequent agent will choose $A$, so choosing $A$ is optimal.

        Note that the solution from backward induction does include the off-equilibrium path action. We can use this rationality requirement to show that the first agent will choose $A$. The subsequent agents will also follow in choosing $A$.

        In the case where $n$ is odd, there is no need for tie-breaking, making $\lambda$ irrelevant. We similarly, have that, by denoting $a$ as total number of agents who has proposed $A$ so far, the equilibrium action for the current agent is to choose $B$ if and only if $a \le \frac{n-1}{2}$. The induction process is similar.
    \end{proof}

    We also have a converse result
    \begin{theorem}
        If $\lambda <0$ and $\epsilon >0$, then under regularities $\abs{\lambda} < \frac{1}{n}$, $\abs{\epsilon} < \frac{1}{n^2}$, $\abs{\Theta}=2$, and $T >\frac{\max_{\theta'} \ln \Pi(\{\theta'\}) - \min_{\theta'} \ln \Pi(\{\theta'\})}{\abs{\lambda}}$, then we have that
        \begin{itemize}
            \item If $n$ is odd, then, in equilibrium, every ASI agent chooses
            \begin{align*}
                \hat{\theta}_i = \theta.
            \end{align*}
            \item If $n$ is even, then, in equilibrium, every ASI agent chooses
            \begin{align*}
                \hat{\theta}_i \ne \theta.
            \end{align*}
        \end{itemize}
    \end{theorem}
    \begin{proof}
        The proof is the same as that of the theorem~\ref{thm:1}.
    \end{proof}

    These two theorems suggest that the nature of ASI agents and the problem of interest is important for the design of the debate in how many agents we should deploy.
    
\section{Connection to AI Debate Literature}

    The paper~\cite{irving2018ai} introduces the notion of AI debates. It briefly mentions the zero-sum aspect of the game but does not formalize nor analyze the game. \cite{kovavrik2019truth} attempts to formalize into a game but with reduced form result where the designer can choose the utility function of the agents.
    
    Other papers require the observation of the unobserved state for its game formulation~\cite{brown2025avoiding}, or creates a fictitious game merely for analysis and training.\cite{brown2024scalable,chen2024playing,brown2026debate} In this paper, we formally analyze the behavior of the game created by a debate system where the agents are superintelligences.

\bibliographystyle{alpha}
\bibliography{egbib}

@article{irving2018ai,
  title={AI safety via debate},
  author={Irving, Geoffrey and Christiano, Paul and Amodei, Dario},
  journal={arXiv preprint arXiv:1805.00899},
  year={2018}
}

@article{kim2024road,
  title={The Road to Artificial SuperIntelligence: A Comprehensive Survey of Superalignment},
  author={Kim, HyunJin and Yi, Xiaoyuan and Yao, Jing and Lian, Jianxun and Huang, Muhua and Duan, Shitong and Bak, JinYeong and Xie, Xing},
  journal={arXiv preprint arXiv:2412.16468},
  year={2024}
}

@article{barrett2017model,
  title={A model of pathways to artificial superintelligence catastrophe for risk and decision analysis},
  author={Barrett, Anthony M and Baum, Seth D},
  journal={Journal of Experimental \& Theoretical Artificial Intelligence},
  volume={29},
  number={2},
  pages={397--414},
  year={2017},
  publisher={Taylor \& Francis}
}

@inproceedings{brown2024scalable,
  title={Scalable AI Safety via Doubly-Efficient Debate},
  author={Brown-Cohen, Jonah and Irving, Geoffrey and Piliouras, Georgios},
  booktitle={International Conference on Machine Learning},
  pages={4585--4602},
  year={2024},
  organization={PMLR}
}

@article{kovavrik2019truth,
  title={(When) Is Truth-telling Favored in AI Debate?},
  author={Kova{\v{r}}{\'\i}k, Vojt{\v{e}}ch and Carey, Ryan},
  journal={arXiv preprint arXiv:1911.04266},
  year={2019}
}

@article{brown2025avoiding,
  title={Avoiding obfuscation with prover-estimator debate},
  author={Brown-Cohen, Jonah and Irving, Geoffrey and Piliouras, Georgios},
  journal={arXiv preprint arXiv:2506.13609},
  year={2025}
}

@inproceedings{chen2024playing,
  title={Playing Large Games with Oracles and AI Debate},
  author={Chen, Xinyi and Chen, Angelica and Foster, Dean and Hazan, Elad},
  booktitle={Agentic Markets Workshop at ICML 2024},
  year={2024}
}

@article{brown2026debate,
  title={Debate is efficient with your time},
  author={Brown-Cohen, Jonah and Irving, Geoffrey and Marshall, Simon C and Newman, Ilan and Piliouras, Georgios and Szegedy, Mario},
  journal={arXiv preprint arXiv:2602.08630},
  year={2026}
}

\end{document}